\documentclass[runningheads]{llncs}
\usepackage{graphicx}
\usepackage{amssymb}
\usepackage{amsmath}
\usepackage{cite}
\begin{document}
\title{A Siamese Neural Network with Modified Distance Loss For Transfer Learning in Speech Emotion Recognition}
\titlerunning{Siamese Network with Modified Distance Loss in Transfer Learning}
%
\author{Kexin Feng \and Theodora Chaspari}
\authorrunning{Feng et al.}
%

\institute{HUman Bio-Behavioral Signals (HUBBS) Lab \\
Department of Computer Science and Engineering \\
Texas A\&M University \\
\email{\{kexin0814, chaspari\}@tamu.edu}\\}

\maketitle              
\begin{abstract}
Automatic emotion recognition plays a significant role in the process of human computer interaction and the design of Internet of Things (IOT) technologies. Yet, a common problem in emotion recognition systems lies in the scarcity of reliable labels. By modelling pairwise differences between samples of interest, a Siamese network can help to mitigate this challenge since it requires fewer samples than traditional deep learning methods. In this paper, we propose a distance loss, which can be applied on the Siamese network fine-tuning, by optimizing the model based on the relevant distance between same and different class pairs. Our system uses samples from the source data to pre-train the weights of proposed Siamese neural network, which are fine-tuned based on the target data. We present an emotion recognition task that uses speech, since it is one of the most ubiquitous and frequently used bio-behavioral signals. Our target data comes from the RAVDESS dataset, while the CREMA-D and eNTERFACE’05 are used as source data, respectively. Our results indicate that the proposed distance loss is able to greatly benefit the fine-tuning process of Siamese network. Also, the selection of source data has more effect on the Siamese network performance compared to the number of frozen layers. These suggest the great potential of applying the Siamese network and modelling pairwise differences in the field of transfer learning for automatic emotion recognition.

\end{abstract}
\section{Introduction}
Automatic emotion recognition refers to identifying emotions using various human-related signals such as facial expression, physiological signals, and speech~\cite{picard2000affective}. It can potentially benefit with many applications related to human computer interaction, health informatics, or even the design of smart cities and communities. Among these signals, speech data is largely explored due to its relatively higher availability and ease of collection. Acquiring reliable annotation for such large amounts of audio clips can be extremely hard to obtain, providing a significant impediment for the reliable training of emotion recognition systems.

A great number of machine learning approaches have been proposed to address this challenge, and transfer learning was shown to be one of the most promising directions. Transfer learning methods such as fine-tuning~\cite{badshah2017speech} make use of a well-trained model on another emotion dataset. Also, progressive neural networks (PNN)~\cite{gideon2017progressive}, which are less forgetful when applied to target, have been proposed and obtained good performance in leveraging knowledge between various conditions. However, these methods might be less effective when there are very limited number of data in target domain, preventing adequate training of the corresponding machine learning models. 

Modelling the pairwise differences between samples of interest (e.g., through Siamese networks~\cite{koch2015siamese}) could be a potential solution to address small amounts of labelled target data in various applications. In this paper, we propose the use of Siamese network structure for the task of transfer learning in speech emotion recognition, trained using the fine-tuning method, and further optimized using a distance loss that incorporates relative distance among pairs. A publicly available dataset, RAVDESS~\cite{livingstone2012ravdess}, was used as target data due to its relatively small number of speakers and sample size. Two other emotional datasets, CREMA-D~\cite{cao2014crema} and eNTERFACE’05~\cite{martin2006enterface}, are used as source data to compare the influence of different source domains. Our results indicate that the selection of source data can have a significant impact on the Siamese network fine-tuning, and our distance loss can significantly benefit the Siamese network fine-tuning process, which yields an improvement of up to 7\% compared to fine-tuning the Siamese network without the proposed distance loss. 

\section{Previous Work}
The Siamese network is a type of neural network which takes a pair of data samples as an input, and decides whether the corresponding samples belong to the same or different classes. This network structure was first introduced by Bromley et al. for the task of signature verification by comparing whether the two signatures are from the same person~\cite{bromley1994signature}. In Siamese network, each data sample in a pair is the input to the network, then the network outputs an extracted feature vector. The $l2$-norm between the two extracted feature vectors is further calculated based on the two extracted features. If the $l2$-norm distance is less than a certain threshold point, the model will decide the samples in this pair are from the same class, otherwise the samples belong to different classes.

Researchers further optimized the initially proposed structure of the Siamese network by considering the data distribution within the classes of interest~\cite{koch2015siamese}. Instead of calculating the $l2$-norm, the $l1$-norm vector was calculated, and was followed by fully connected layers with sigmoid activation to allow the model learn the relations between distance and classes.

The Siamese network has been applied to emotion-related tasks with a promising performance~\cite{lian2018speech,huang2018speech,sabri2018facial}. However, to the best of our knowledge, this is the first time the Siamese network was applied and further optimized for emotion-related transfer learning task.

\section{Data Description}
The target data comes from the RAVDESS dataset~\cite{livingstone2012ravdess}, which includes 47 minutes of audio from 24 actors, and each audio sample is further labeled by 247 annotators. The source data comes from two different publicly available datasets: the eNTERFACE’05~\cite{martin2006enterface} and CREMA-D~\cite{cao2014crema}. The eNTERFACE’05 contains 45 minutes of data from 42 participants, who were asked to express their emotions in scripted sentences after listening to specific stories. The CREMA-D dataset contains 165 minutes of audio data from 91 actors, who were asked to perform until they get approved by a director and each audio clip is further labeled by human annotators.

Speech samples depicting four common emotions (anger, happiness, sadness, and fear) across all datasets were selected and processed using openSMILE toolkit~\cite{eyben2010opensmile}. A 64-dimensional feature set, part of the INTERSPEECH’09 emotion challenge feature set~\cite{schuller2009interspeech}, is extracted from each audio segment which includes the mean and standard deviations of speech intensity, zero-crossing rate, voice probability, fundamental frequency, and the first 12 Mel-frequency cepstral coefficient (MFCC). The first order derivative of each of these features is also computed.

\section{Methodology}
In this section, we plan to discuss our three baseline methods: in-domain training, out of domain training, and Siamese network fine-tuning. Then we will introduce the proposed distance loss, and how this loss is applied for the Siamese network. All the methods are based on an optimized Siamese network structure, which contains 64, 32, and 16 nodes with ReLU activation for the first three feature extractor layers, followed by a 16-node decision-based layer. During the training process, the cross entropy loss is calculated and used to update the weights of the model. After the Siamese network is trained, the test data will be compared with all the available labeled target data, and a final classification decision will be generated based on the similarity of log-sum with each class. The unweighted average recall (UAR) will be used to evaluate the performance for each method. 

\begin{figure}[htb]
  \includegraphics[width=1\linewidth]{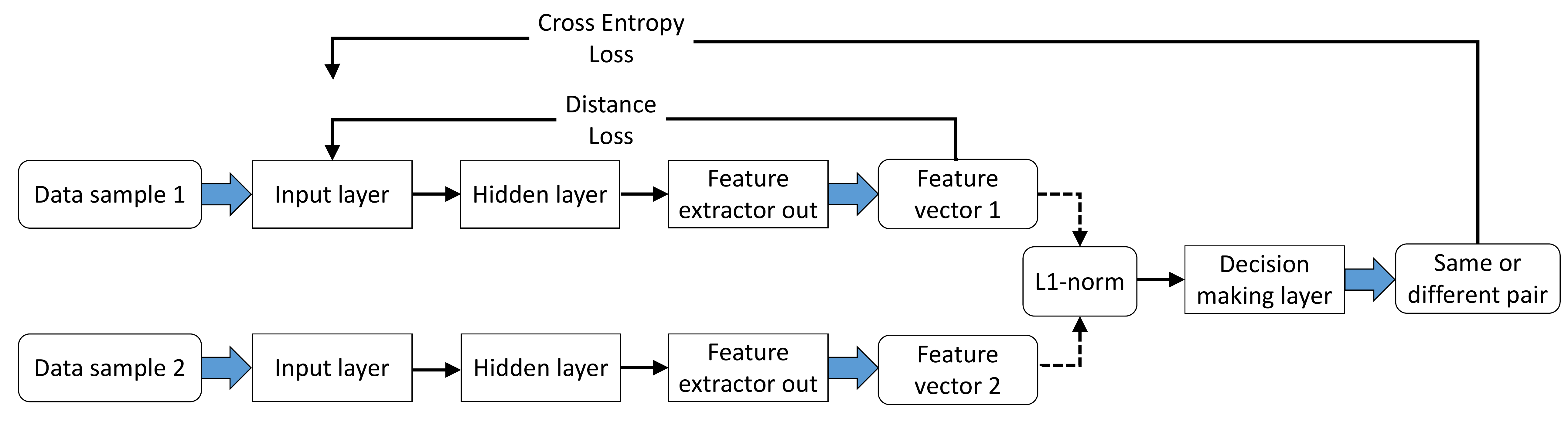}
  \caption{Schematic representation of the proposed Siamese network fine-tuning with modified distance loss.}
  \label{distance_loss}
\end{figure}

\subsection{Baseline}
In order to assess whether the proposed loss will benefit the knowledge transfer efficiency when limited data sample is available, we proposed three baseline methods. The first baseline performs an out-of-domain training (OODT). A Siamese network is trained on all the source data, and tested on the target dataset without any adaptation. For this baseline, the data samples from the source are used to determine the final classes, since no labeled target data is available.

The second baseline is an in-domain training (IDT), which is trained using sufficient amount of target data. A leave-one-subject-out (LOSO) cross-validation is performed. More specifically, samples from a given speaker are used for testing, and all other data are used for training of the Siamese network. This process is repeated until all the speakers have been used for testing. This baseline serves as an upper limit of potential knowledge transfer.

The third baseline is the traditional fine-tuning method in the field of transfer learning, to further assess the benefit provided by our proposed distance loss. The models trained in OODT are used for fine-tuning. One random data sample for each emotion from random 2 speakers are selected as labeled data. As a result, a total number of 8 samples (2 speakers $\times$ 4 emotions $\times$ 1 sample/emotion/speaker) are used for fine-tuning process and the remaining data are used as testing set. This process is repeated 10 times to increase the robustness of the results. A different number of frozen layers (none, first layer, or first two layers), as well as the different number of speakers in target data (2, 4, $\ldots$ , 18, 20) are tested to evaluate in detail the knowledge transfer efficiency. 

\subsection{Siamese NN fine-tuning with modified loss}
Inspired by the fact that the traditional fine-tuning method fails to efficiently leverage the knowledge when limited target data is available, we designed a distance loss to maximize the distance difference between pairs from the same class and pairs from the different class. Let $\mathcal{X}_s$ be the set of $s$ pairs with the same class in a batch, and $\mathcal{X}_d$ is the set of $d$ pairs with the different class. Let $g_\mathbf{W}(\mathbf{x})$ be the function parameterized by the weights $\mathbf{W}$ of the Siamese network that performs the transformation between the original input $\mathbf{x}$ and the extracted feature vector. The parameter $\mathbf{W}$ is learned by minimizing the average relative distance between pairs $\mathcal{X}_s$ of the same class and maximizing the distance between pairs $\mathcal{X}_d$ of different classes:


\begin{equation}
\mathbf{W^*} = arg\min_{\mathbf{W}}\frac{
\frac{1}{\|\mathcal{X}_d\|_0}\sum_{\mathbf{x},\mathbf{x^{'}} \in{\mathcal{X}_d}} \lVert\mathbf{g(\mathbf{x}), g(x^{'})}\rVert _{2} + \frac{1}{\|\mathcal{X}_s\|_0}\sum_{\mathbf{x},\mathbf{x^{'}} \in{X_s}} \lVert\mathbf{g(x), g(x^{'})}\rVert _{2}} {\frac{1}{\|\mathcal{X}_d|_0}\sum_{\mathbf{x},\mathbf{x^{'}} \in{X_d}} \lVert\mathbf{g(x), g(x^{'})}\rVert _{2} - \frac{1}{\|\mathcal{X}_s\|_0}\sum_{\mathbf{x},\mathbf{x^{'}} \in{X_s}} \lVert\mathbf{g(x), g(x^{'})}\rVert _{2}}
\label{equation}
\end{equation}

As indicated in the previous description, this loss is only applied to the feature extraction layers at the end of a batch in the Siamese network fine-tuning process to minimize the possible influence to the decision making process (Figure~\ref{distance_loss}). Besides this loss, the other part for this method remains the same compared to the fine-tuning baseline method.

\begin{table}[t]
\caption{Unweighted average recall (UAR\%) of the out-of-domain training (OODT), in-domain training (IDT), and the best results obtained among the different number of frozen layers and adopted speakers using Siamese NN fine-tuning with / without modified loss.}
\setlength{\tabcolsep}{3.5mm}{
\begin{tabular}{ccc}
\hline
source & eNTERFACE’05 & CREMA-D \\ \hline
OODT & 32.8 & 29.3 \\
IDT & 50.0 & 50.0 \\
Siamese NN fine-tuning & 32.9 & 37.8 \\
\begin{tabular}[c]{@{}c@{}}Siamese NN fine-tuning with modified loss\end{tabular} & 39.9 & 43.4 \\ \hline
\end{tabular}
}
\label{overall}
\end{table}

\section{Results}
In an effort to discuss the domain difference between the different speech emotion datasets, we first examine the unweighted average recall (UAR) for four emotion classification task on OODT baseline. This yields a 32.8\% UAR on the RAVDESS dataset when using eNTERFACE’05 as source and 29.3\% when using CREMA-D as source. We then examines the upper limit of classification performance using the proposed feature and model structure, and obtained a UAR at 50.0\% with in-domain training.

\begin{figure*}[ht]
\begin{minipage}[t]{0.32\linewidth}
  \centering
  \centerline{\includegraphics[trim = 5mm 0mm 0mm 0mm, clip=true, scale=0.28]{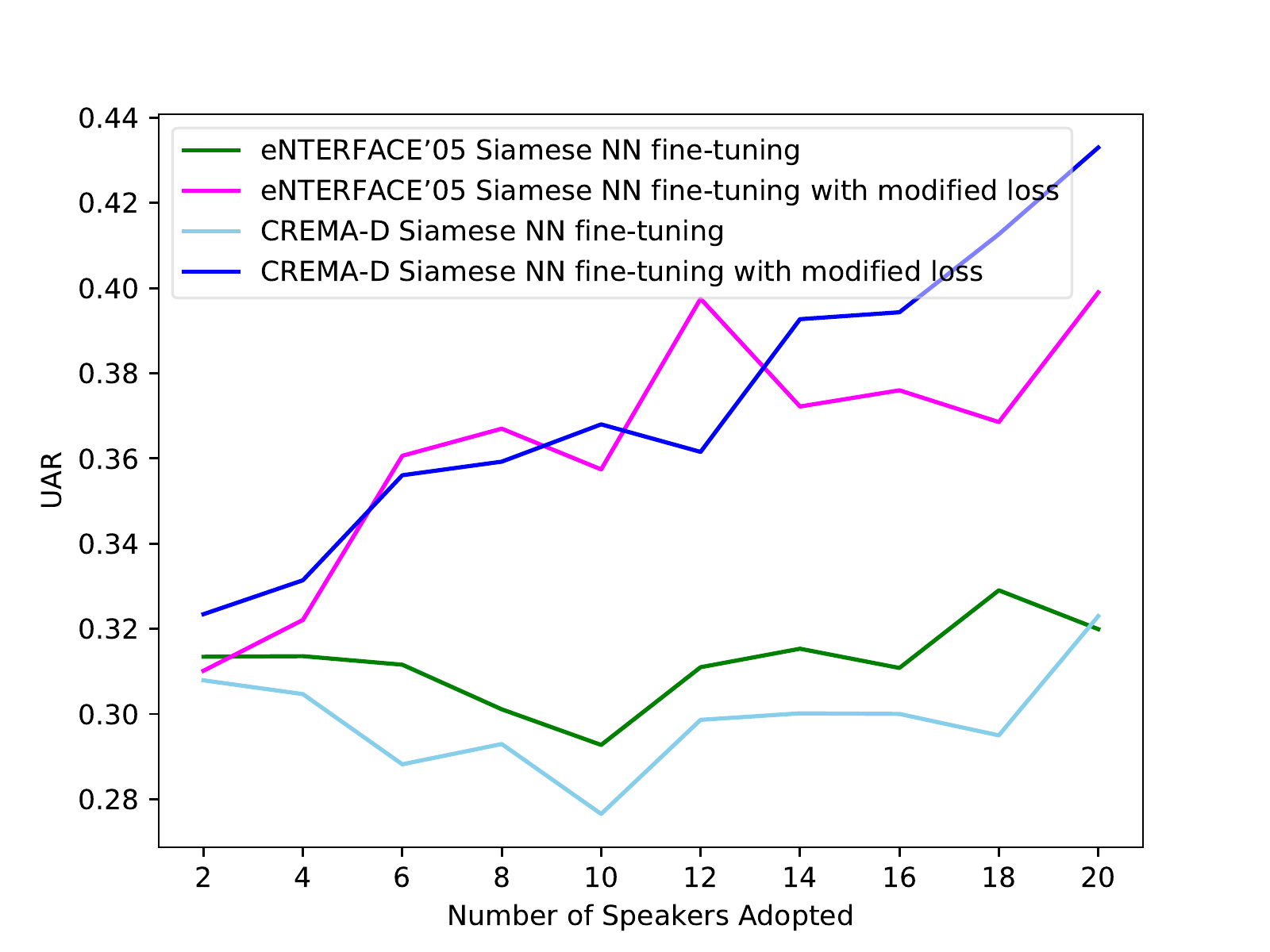}}
  \centerline{(a) No frozen layers.}\medskip
\end{minipage}
\begin{minipage}[t]{0.32\linewidth}
  \centering
  \centerline{\includegraphics[trim = 5mm 0mm 0mm 0mm, clip=true, scale=0.28]{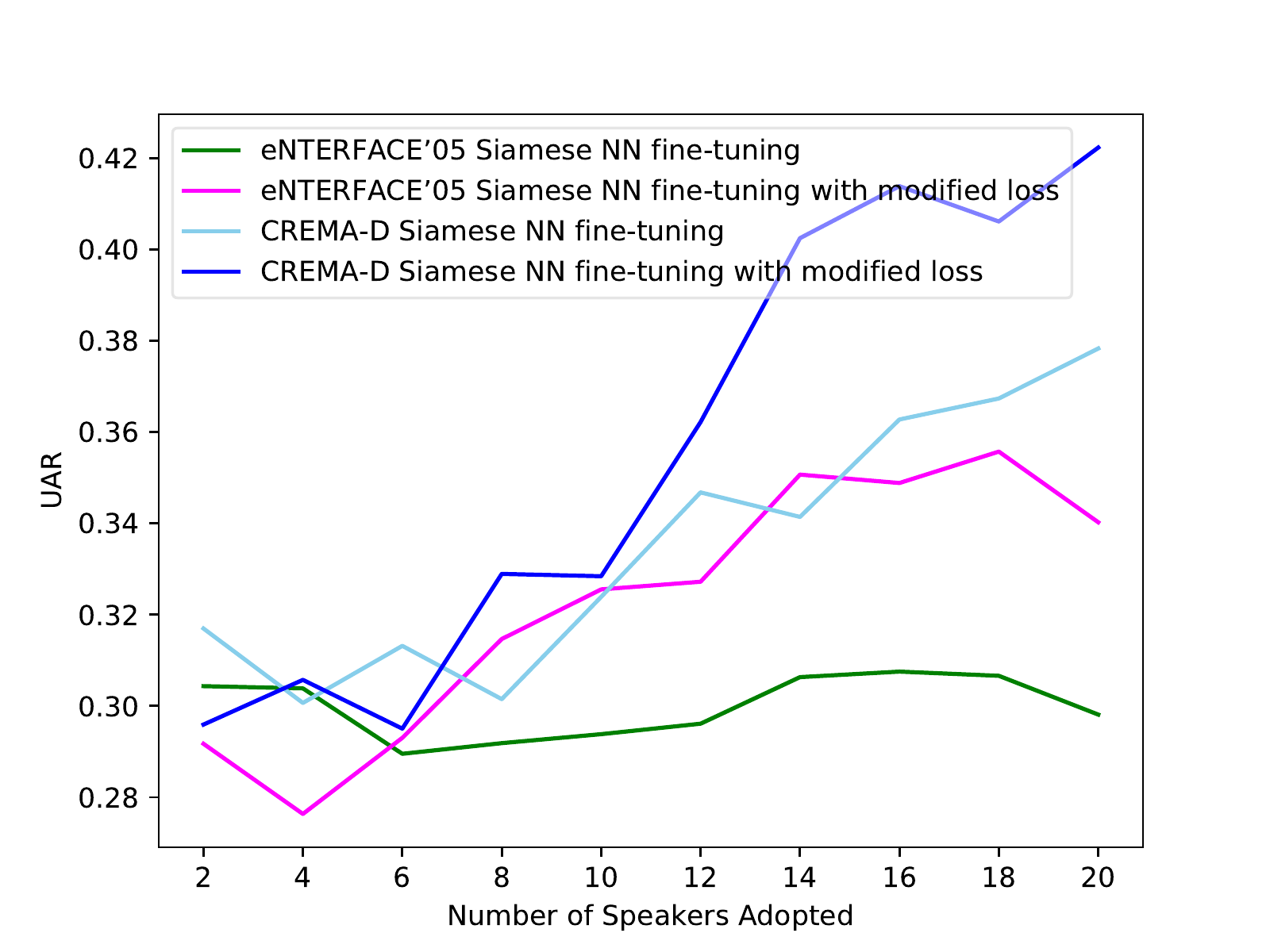}}
  \centerline{(b) First layer is frozen.}\medskip
\end{minipage}
\begin{minipage}[t]{0.32\linewidth}
  \centering
  \centerline{\includegraphics[trim = 5mm 0mm 0mm 0mm, clip=true, scale=0.28]{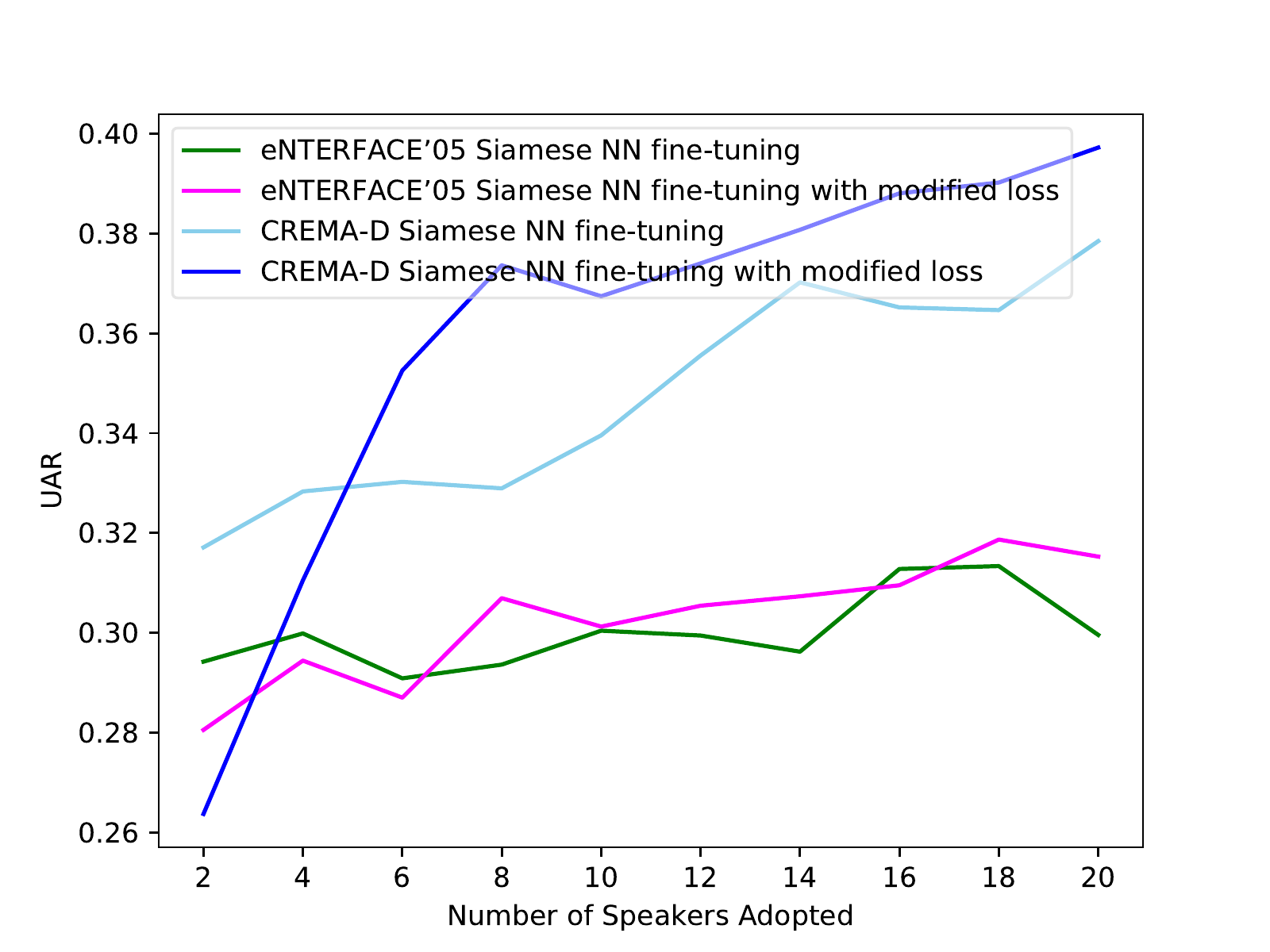}}
  \centerline{(c) First two layers are frozen.}\medskip
\end{minipage}
\caption{Unweighted average recall (UAR) for freeze different number of layers and different number of speakers adopted in fine-tuning process with / without distance loss.}
\label{fig:res}
\end{figure*}

The fine-tuning of the Siamese neural network without the proposed loss is used to illustrated the knowledge transfer efficiency. This approach resulted in a minor improvement when using eNTERFACE’05 as source data at 32.9\%, and a relatively large improvement when using CREMA-D as source data at 37.8\%. We finally added our proposed distance loss in the fine-tuning process, and obtained a significant improvement of 39.9\% using eNTERFACE’05 and 43.3\% using CREMA-D. A comparison of the performance of different methods can be found at Table~\ref{overall}. Our results also indicate that compared with different number of frozen layers, different source data may has a more important role for Siamese network fine-tuning (Figure~\ref{fig:res}).

\section{Discussion}
Our results indicate that including 3-5 speakers in the Siamese NN fine-tuning with modified loss may result in the best trade-off between performance and the number of data samples used for training. This can be potentially explained by the fact that this number of speakers might not be enough to capture the distribution of the target dataset. Performance is degraded when including a smaller number of speakers in the target data. Also, if the data distribution includes a lot of variability, Siamese NN with modified loss will be greatly restricted, since pairwise differences are less likely to express the class information.

There has not been a lot of research relevant to few-shot emotion recognition with speech signals, where data from only a few number of speakers is included in the target data. The most comparable results are from Gideon et al., in which the proposed approach with progressive neural networks achieved a limited improvement (around 3\%) compared to traditional fine-tuning (\cite{gideon2017progressive}). Our proposed distance loss is able to bring a relative larger improvement at about 7\% compared to fine-tuning, without significant increase in the computational cost. Even though Siamese NN fine-tuning with modified loss still fails to outperform the in-domain training, it shows great potential. We will attempt to modify the proposed methodology and explore whether it can reach in-domain performance as part of our future work. 

\section{Conclusions}
We propose a distance loss which is based on the relative distance between same and different class pairs. Such loss can increase the upper limit and increase knowledge transfer efficiency when very limited target data is available. Our results also indicate that the selection of source data plays a more important role than the number of frozen layers. Findings of this work can be applied on other tasks using Siamese network, fine-tuning, or few-shot learning. The application of distance between different emotion classes can be a fundamental step in understanding the difference between emotion categories.

As part of our future work, we plan to explore the usage of pairs in addressing the domain mismatch, and mitigate the influence of different source data in the process of knowledge transfer. We will further perform the multi-source transfer learning by using pairwise information to select proper source data from each dataset. Finally, we plan to understand the distance between different emotions with the help of the pairing information.

\bibliographystyle{splncs04}
\bibliography{refs.bib}

\end{document}